\documentclass[sigconf]{acmart}
\AtBeginDocument{%
  \providecommand\BibTeX{{%
    Bib\TeX}}}

\setcopyright{acmlicensed}
\copyrightyear{2018}
\acmYear{2018}
\acmDOI{XXXXXXX.XXXXXXX}
\acmConference[Conference acronym 'XX]{Make sure to enter the correct
  conference title from your rights confirmation email}{June 03--05,
  2018}{Woodstock, NY}
\acmISBN{978-1-4503-XXXX-X/2018/06}

\usepackage{algpseudocode}
\usepackage{graphicx}
\usepackage[normalem]{ulem}  
\usepackage{xspace}
\usepackage{textcomp}
\usepackage{xcolor}
\usepackage{adjustbox}  
\usepackage{algorithm}
\usepackage{multirow}
\usepackage{makecell}
\usepackage{booktabs}
\usepackage{color}
\usepackage{subcaption}
\usepackage[para]{footmisc} 
\usepackage{tabularx}
\usepackage{booktabs}
\usepackage{colortbl}
\usepackage{multirow}
\usepackage{arydshln}

\usepackage{enumitem}

\definecolor{lightorange}{RGB}{255, 242, 204}

\newcommand{\cmt}[1]{}

\AtBeginDocument{%
  \providecommand\BibTeX{{%
    Bib\TeX}}}
\renewcommand\footnotetextcopyrightpermission[1]{}

\definecolor{cbad}{HTML}{F2F4F7}

\definecolor{clow}{HTML}{FFF9E6}   
\definecolor{cmed}{HTML}{FFECB3}   
\definecolor{chigh}{HTML}{FFD699} 

\newcommand{\cb}{\cellcolor{cbad}}
\newcommand{\cl}{\cellcolor{clow}}
\newcommand{\cm}{\cellcolor{cmed}}
\newcommand{\ch}{\cellcolor{chigh}}

\begin{document}

\title{Exploiting Function-Family Structure in Analog Circuit Optimization}


\author{Zhuohua Liu\textsuperscript{1}, Kaiqi Huang\textsuperscript{2}, Qinxin Mei\textsuperscript{2}, Yuanqi Hu\textsuperscript{1*}, Wei W. Xing\textsuperscript{3*}}
\affiliation{%
  \institution{\textsuperscript{1}School of Integrated Circuit Science and Engineering, Beihang University \\  \textsuperscript{2}School of Mechatronic Control Engineering, Shenzhen University \\ \textsuperscript{3}School of Mathematical and Physical Science, University of Sheffield}
  \city{}
  \country{}
  }
\email{ZhuohuaLiu@buaa.edu.cn}






\begin{abstract}

Analog circuit optimization is typically framed as black-box search over arbitrary smooth functions, yet device physics constrains performance mappings to structured families: exponential device laws, rational transfer functions, and regime-dependent dynamics. Off-the-shelf Gaussian-process surrogates impose globally smooth, stationary priors that are misaligned with these regime-switching primitives and can severely misfit highly nonlinear circuits at realistic sample sizes (50–100 evaluations). We demonstrate that pre-trained tabular models encoding these primitives enable reliable optimization without per-circuit engineering. Circuit Prior Network (CPN) combines a tabular foundation model (TabPFN v2) with Direct Expected Improvement (DEI), computing expected improvement exactly under discrete posteriors rather than Gaussian approximations. Across 6 circuits and 25 baselines, structure-matched priors achieve $R^2 \approx 0.99$ in small-sample regimes where GP-Matérn attains only $R^2 = 0.16$ on Bandgap, deliver $1.05$--$3.81\times$ higher FoM with $3.34$--$11.89\times$ fewer iterations, and suggest a shift from hand-crafting models as priors toward systematic physics-informed structure identification. Our code will be made publicly available upon paper acceptance.

\end{abstract}



\keywords{Transistor Sizing, Analog Circuit Design, Bayesian Optimization, Tabular Foundation Models, Function-Family Priors}



\AtBeginDocument{%
  \providecommand\BibTeX{{%
    Bib\TeX}}}
\settopmatter{printacmref=false}
\pagestyle{plain} 
\renewcommand\footnotetextcopyrightpermission[1]{}

\maketitle

\section{Introduction}

Analog circuit design represents a \$100B+ global semiconductor market, with transistor sizing consuming 30--50\% of design time~\cite{poddar2025insight}. The challenge is severe: SPICE simulations require seconds to hours per evaluation, design spaces span tens to over fifty variables, and tight specifications create narrow feasible regions where minor violations are catastrophic. Sample efficiency is critical because practical budgets typically allow only a few hundred evaluations.

Yet circuit performance mappings are far from arbitrary. Device equations impose strong constraints: weak inversion yields exponential $I_D \propto \exp(V_{GS}/nV_T)$, strong inversion follows power-law $I_D \propto (W/L)(V_{GS} - V_{th})^{\alpha}$, and $g_m/I_D$ saturates quasi-monotonically~\cite{enz1995analytical}. Threshold conditions create sharp regime boundaries at $V_{th}$ and $V_{DS, sat}$. Small-signal analysis yields rational transfer functions $H(s) = N(s)/D(s)$ whose poles and zeros move continuously with geometry and bias~\cite{gray2009analysis}. Gain, bandwidth, and phase margin couple through pole-zero spacings. Large-signal dynamics add exponential RC transients, slew-limited ramps, and regime-dependent ODEs. These are not arbitrary smooth functions but compositions of a finite vocabulary: exponential device laws, rational network responses, and piecewise regime-switching behavior. Performance landscapes form a \textit{structured function family} constrained by physics.

Most current Bayesian optimization methods for analog circuit sizing ignore this structure and employ Gaussian-process surrogates with RBF or Mat\'ern kernels~\cite{zhang2021efficient,xing2024kato,li2025mario}, which impose globally smooth, stationary priors that do not encode the regime-switching behavior implied by circuit primitives. At realistic sample sizes of 50--100 evaluations, this mismatch limits surrogate fidelity: stationary kernels with a single characteristic length-scale cannot explicitly capture exponential device laws, nonlinear regime transitions, or sharp operating-region boundaries, leading to uncertain and poorly calibrated predictions. In these low-data regimes, the surrogate often fails to recover useful structure, yielding near-zero or even negative $R^2$ on held-out points. This behavior is not merely due to hyperparameter choices but reflects an inductive-bias mismatch between globally smooth, stationary priors and the underlying circuit primitives.

If performance mappings live in a known function family, we should encode that structure as an inductive bias rather than fit arbitrary smooth functions. This reframes optimization from generic function approximation to identifying which family member matches observations. Prior-data Fitted Networks (PFNs)~\cite{muller2022transformers,hollmann2025accurate} provide a validation path: meta-trained on 100K+ synthetic regression tasks, TabPFN v2~\cite{hollmann2025accurate} performs Bayesian inference in context without per-task training. Its training distribution explicitly includes exponential components, rational responses, and regime-dependent boundaries—structures that directly match the governing transistor-level circuits. 
When applied to analog sizing, such physics-aligned priors yield substantially higher surrogate accuracy in low-sample regimes, providing support for the function-family hypothesis.

We develop this insight into Circuit Prior Network (CPN), which treats analog sizing as inference over pre-trained function-family priors. CPN reduces per-circuit surrogate engineering by replacing iterative GP kernel selection with TabPFN v2 models that encode circuit primitives. We introduce Direct Expected Improvement (DEI), which computes expected improvement exactly under the surrogate's discrete posterior instead of a Gaussian approximation, making it well-suited to skewed or multimodal posteriors near regime boundaries.
Our contributions are:
\begin{enumerate}

    \item We introduce the CPN, the first framework to formulate analog sizing as inference over function-family priors derived from circuit primitives. This function-family perspective bridges circuit theory and foundation models, enabling a reusable prior across heterogeneous designs.

    \item We propose DEI, a distribution-consistent acquisition rule that evaluates improvement exactly under TabPFN v2's discrete posterior.

    \item Through experiments, we compare sizing formulations—direct vs.\ decomposed FoM prediction and direct FoM maximization vs.\ constrained optimization—and show that direct FoM modeling yields faster convergence and better final designs.

    \item Across six circuits and 25 baselines, structure-matched priors achieve up to $R^2 \approx 0.99$ in small-sample regression, deliver up to $3.62\times$ gains in normalized constrained objectives, and failure analysis confirms structure matching as the key performance driver.
    
\end{enumerate}

\section{Background}
\label{sec:background}

\subsection{Problem Formulation}
\label{sec:problem_formulation}
Transistor sizing seeks to optimize circuit performance while satisfying design specifications. We formulate this as constrained black-box optimization:
\begin{equation}
    \max_{\mathbf{x} \in \mathcal{X}} f_0(\mathbf{x}) \quad \text{s.t.} \quad f_i(\mathbf{x}) \geq C_i, \; i = 1, \dots, N_c,
\end{equation}
where $\mathbf{x} \in \mathbb{R}^D$ represents design variables (transistor widths, lengths, multiplicities, bias voltages, and currents), $f_i(\mathbf{x})$ denotes the $i$-th performance metric obtained via SPICE simulation, and $C_i$ is the corresponding constraint threshold. Evaluating $f_i(\mathbf{x})$ requires expensive simulation, making sample efficiency critical.
We simplify optimization by defining a scalar Figure of Merit (FoM) aggregating normalized specifications:
\begin{equation}
    \label{eq:fom}
    \text{FoM}(\mathbf{x}) = \sum_{i=1}^{N} s_i(\mathbf{x}),
\end{equation}
where $s_i(\mathbf{x})$ quantifies how well the $i$-th specification is satisfied. For hard constraints, $s_i(\mathbf{x}) = \min(f_i(\mathbf{x})/C_i, 1)$ or $\min(C_i/f_i(\mathbf{x}), 1)$ depending on whether the metric is maximized or minimized. 
For optimization targets, $s_i(\mathbf{x}) = f_i(\mathbf{x})/C_i$ or $C_i/f_i(\mathbf{x})$ once all hard constraints are met. 
Maximizing Eq.~\ref{eq:fom} therefore corresponds to satisfying all specifications (each contributing 1) and then improving performance beyond the required thresholds.

Note that some circuits use domain-specific "FoM" definitions (e.g., Gm). These differ from the scalar FoM in Eq.~\ref{eq:fom}, which serves as our optimization objective.

\subsection{Circuit Response Families}

Analog circuit behaviors are far from arbitrary black-box functions. 
Device laws introduce exponential and power-law structure, along with abrupt regime boundaries when transistors transition between cutoff, triode, and saturation. 
At the network level, small-signal models yield low-order rational transfer functions whose poles and zeros vary smoothly with geometry and bias conditions. 
Large-signal operation further induces piecewise dynamics such as exponential charging, slew-limited ramps, and mode switching near critical operating points.
These behaviors can be viewed as compositions of a structured function family:
\begin{equation}
\mathcal{F}_{\text{circuit}} = \text{span}\left\{\exp(\cdot),\; (\cdot)^\alpha,\; \frac{p(s)}{q(s)},\; \mathbb{I}_{\text{region}}(\cdot)\right\},
\end{equation}
capturing exponential device responses, power-law saturation, rational network transfer characteristics, and indicator-based regime transitions. 
Circuit performance mappings $\mathbf{x} \mapsto y$ therefore have strong inductive structure and are typically nonstationary, nonlinear, and piecewise smooth.

Standard Gaussian process surrogates rely on stationary smooth kernels such as RBF or Matérn, which do not explicitly encode exponential curvature, rational transfer behavior, or sharp regime boundaries. This mismatch induces overly smooth posteriors that fail to capture the multi-regime behavior common in analog circuits, especially in the small-sample regime.

\subsection{Bayesian Optimization Fundamentals}

Bayesian optimization addresses expensive black-box optimization by constructing a probabilistic surrogate model to guide sequential evaluation. Given observation history $\mathcal{D}_t = \{(\mathbf{x}_i, y_i)\}_{i=1}^t$, a surrogate provides predictive distribution $p(f \mid \mathcal{D}_t)$, typically summarized by mean $\mu_t(\mathbf{x})$ and variance $\sigma_t^2(\mathbf{x})$. An acquisition function $\alpha_t(\mathbf{x})$ balances exploration of uncertain regions and exploitation of promising areas. Expected Improvement is a common choice:
\begin{equation}
\label{eq:ei}
\alpha_{\text{EI}}(\mathbf{x}) = \mathbb{E}\left[\max(f(\mathbf{x}) - f^\star, 0)\right] = (\mu - f^\star)\Phi(z) + \sigma\phi(z),
\end{equation}
where $f^\star = \max_{i \leq t} y_i$ is the current best observation, $z = (\mu - f^\star)/\sigma$, and $\Phi(\cdot)$ and $\phi(\cdot)$ are the standard normal CDF and PDF. The next evaluation point maximizes acquisition:
\begin{equation}
\mathbf{x}_{t+1} = \arg\max_{\mathbf{x} \in \mathcal{X}} \alpha_t(\mathbf{x}).
\end{equation}

Standard BO employs Gaussian process surrogates with stationary kernels (RBF, Matérn), which encode globally smooth priors. At realistic sample sizes of $N < 100$ in high-dimensional design spaces ($D > 20$), kernel misspecification can be severe: stationary kernels do not explicitly encode exponential device laws, sharp regime boundaries, or rational transfer behavior inherent to analog circuits. In practice, this often yields overly smooth posteriors that fail to capture skewed or multimodal structure near operating-point transitions, which in turn can degrade acquisition quality precisely where circuit behavior is most nonlinear.

\vspace{-0.2cm}
\section{Circuit Prior Network Framework}
\label{sec:proposed_method}

Given the function-family structure identified in Section~\ref{sec:background}, we require a surrogate that: (1) encodes exponential, power-law, and rational function families; (2) handles regime boundaries and piecewise behavior; (3) performs reliable inference at 50--100 samples without per-task engineering. We instantiate this via TabPFN v2 and develop a distribution-consistent acquisition strategy.

\subsection{TabPFN v2 as Function-Family Prior}

\noindent\textbf{Amortized Bayesian inference.} TabPFN v2~\cite{hollmann2025accurate} performs amortized Bayesian inference by meta-training a Transformer on large collections of synthetic tabular prediction tasks. Each task is generated by sampling a latent function from a broad generative prior, producing diverse nonlinear relationships. After meta-training, the model maps a context $\mathcal{D}_t=\{(\mathbf{x}_i,y_i)\}_{i=1}^t$ and query $\mathbf{x}$ to a predictive distribution in a single forward pass, requiring no gradient updates or per-task hyperparameter tuning. This avoids the repeated kernel adaptation and $O(N^3)$ scaling associated with Gaussian-process surrogates during BO.

\noindent\textbf{Function-family alignment.} Although no circuit data appear in pretraining, TabPFN v2’s synthetic task generator produces nonlinear behaviors that overlap with the primitive function families in Section~\ref{sec:background}. Its tasks combine neural activations (sigmoid, ReLU, sine), multiplicative interactions, discretization, clipping, and piecewise transformations, yielding functions with exponential-like growth, power-law behavior, rational-like interactions, and regime-dependent transitions. This broad structural prior provides a closer inductive match to transistor- and network-level response surfaces than stationary GP kernels, explaining its stability in the low-sample regime.

\noindent\textbf{Discrete predictive posterior.} For regression, TabPFN v2 outputs a piecewise-constant approximation to the predictive posterior:
\begin{equation}
P(y \mid \mathbf{x}, \mathcal{D}_t)=\{(c_k, p_k(\mathbf{x}))\}_{k=1}^K,
\end{equation}
where $\{c_k\}$ are fixed value bins and $p_k(\mathbf{x})$ is the predicted probability mass at bin $k$; $K$ is the number of bins. This form yields standard moments,
\begin{equation}
\mu(\mathbf{x}) = \sum_{k=1}^K c_k p_k(\mathbf{x}), \qquad
\sigma^2(\mathbf{x}) = \sum_{k=1}^K (c_k-\mu(\mathbf{x}))^2 p_k(\mathbf{x}),
\end{equation}
while preserving full distributional shape, including skewness and multimodality—structures that Gaussian-posterior GPs cannot express. A single forward pass computes all $p_k(\mathbf{x})$, making inference cost-effectively constant within the model’s supported sequence length and negligible relative to SPICE simulation time.

\subsection{Direct Expected Improvement Acquisition}

\noindent\textbf{Failure Modes of Moment-Based Acquisition.} Classical Expected Improvement assumes Gaussian posteriors and operates only on mean $\mu(\mathbf{x})$ and variance $\sigma^2(\mathbf{x})$, using the closed-form expression in Eq.~\eqref{eq:ei}. This two-moment summary fails when circuit posteriors deviate from a Gaussian shape. Consider three common cases. First, exponential device laws create asymmetric tails: the true distribution may concentrate 15\% probability mass in a narrow spike above $f^\star$ with a broad tail below. The Gaussian approximation spreads variance symmetrically, underweighting the high-value tail. Second, regime switching produces bimodal distributions with one mode in the feasible region and another in the infeasible region. A Gaussian fit centers between modes, assigning high EI to the boundary where performance is actually poor. Third, saturation boundaries generate skewed distributions with sharp cutoffs. Variance $\sigma^2$ becomes large due to tail spread, but most probability mass lies far from the improvement region.

\noindent\textbf{DEI Formulation.} We compute expected improvement directly from TabPFN’s discrete posterior rather than fitting a Gaussian. Let $\{(c_k, p_k(\mathbf{x}))\}_{k=1}^K$ denote the support points and probabilities of the predicted piece-wise constant distribution. Then
\begin{equation}
\label{eq:dei}
\text{DEI}(\mathbf{x}) = \sum_{k=1}^K \max(0, c_k - f^\star)\, p_k(\mathbf{x}),
\end{equation}
where $f^\star = \max_{(\mathbf{x}_i, y_i)\in\mathcal{D}_t} y_i$ is the best observation. DEI integrates $\max(0, y - f^\star)$ over the PFN’s discrete posterior: bins with $c_k \le f^\star$ contribute zero, and those with $c_k > f^\star$ contribute proportionally to their probability and improvement. This yields EI exactly with respect to the surrogate’s discrete posterior, avoiding the Gaussian or moment-based approximations used in GP-based EI.

DEI has three key properties. First, it is distribution-consistent: skewness, multimodality, and tail mass directly affect the acquisition value. Second, for a Gaussian predictive posterior, refining the discretization makes Eq.~\eqref{eq:dei} converge to the classical closed-form EI, as the sum becomes a Riemann approximation of the EI integral. Third, the cost is $O(K)$ per candidate; with a small fixed $K$ (e.g., $100$), this is effectively the same order as Gaussian EI and negligible compared to TabPFN inference or SPICE evaluation. Empirically, DEI provides the largest gains when the posterior is strongly non-Gaussian—e.g., exponential sensitivities or regime switching—while behaving similarly to classical EI when the posterior is near-Gaussian. Section~\ref{sec:ei_ablation} evaluates this effect across all benchmarks.

\subsection{Multi-Specification Optimization Strategies}
\label{sec:cpn_formulations}

Section~\ref{sec:problem_formulation} formulated sizing as constrained optimization with an aggregated FoM in Eq.~(\ref{eq:fom}). Because TabPFN v2 performs scalar regression ($\mathbf{x}\!\rightarrow\! y$), multi-specification optimization requires choosing how to structure the surrogate task. We examine three options:

\textbf{(i) Constraint-decomposed modeling.} Separate TabPFN v2 instances model the objective and each constraint margin. A problem with $N_c$ constraints needs $N_c{+}1$ surrogates, and the acquisition function must coordinate their predictions to enforce feasibility.

\textbf{(ii) Metric-decomposed FoM modeling.} Independent TabPFN v2 instances predict primitive performance metrics ($N_m$ surrogates), and the FoM is reconstructed via Eq.~(\ref{eq:fom}). This preserves the semantics of individual specifications and exposes metric coupling to the surrogate.

\textbf{(iii) Direct FoM modeling.} A single TabPFN v2 instance predicts the aggregated FoM directly. Specification checking and normalization occur only during data labeling, reducing surrogate learning to a single scalar regression task.

The strategies differ in both data usage and computation. Decomposed methods split the $N$ observations across multiple surrogates, each learning a partial view of the landscape. Direct modeling concentrates all data into one model and requires only a single PFN forward pass per candidate---a major efficiency benefit when evaluating $10^3$--$10^4$ points per BO iteration. Prior work typically fixes one formulation without comparison; here, we treat it as a design choice. As shown in Section~\ref{sec:formulation_ablation}, the strategy alone can change performance by over 20\% and induce $4.3\times$ variation in end-to-end runtime.

Algorithm~\ref{alg:cpn_bo} summarizes CPN. All TabPFN v2 weights remain frozen; only the context $\mathcal{D}_t$ and incumbent $f^\star$ evolve, eliminating per-circuit surrogate engineering such as kernel selection or hyperparameter tuning.

\begin{algorithm}[t]
\caption{CPN Algorithm}
\label{alg:cpn_bo}
\begin{algorithmic}[1]
\Require Pre-trained TabPFN v2 model $q_{\Phi}$, simulator $f(\mathbf{x})$, design space $\mathcal{X}$, budget $T$
\State Initialize context $\mathcal{D}_0 = \{(\mathbf{x}_i, y_i)\}_{i=1}^{5}$ via random sampling
\For{$t = 1$ \textbf{to} $T$}
    \State Randomly sample candidate set $\mathcal{X}_{\text{cand}} \subset \mathcal{X}$
    \ForAll{$\mathbf{x} \in \mathcal{X}_{\text{cand}}$}
        \State Obtain discrete posterior: $P(y \mid \mathbf{x}, \mathcal{D}_{t-1}) = \{(c_k, p_k(\mathbf{x}))\}_{k=1}^K \gets q_{\Phi}(\mathbf{x}, \mathcal{D}_{t-1})$
        \State Compute acquisition: $\text{DEI}(\mathbf{x}) \gets \sum_{k=1}^K \max(0, c_k - f^\star) \, p_k(\mathbf{x})$
    \EndFor
    \State Select next design: $\mathbf{x}_t \gets \arg\max_{\mathbf{x} \in \mathcal{X}_{\text{cand}}} \text{DEI}(\mathbf{x})$
    \State Evaluate via SPICE: $y_t \gets f(\mathbf{x}_t)$
    \State Update context: $\mathcal{D}_t \gets \mathcal{D}_{t-1} \cup \{(\mathbf{x}_t, y_t)\}$
    \State Update incumbent: $f^\star \gets \max_{(\mathbf{x}_i, y_i) \in \mathcal{D}_t} y_i$
\EndFor
\State \Return $\mathbf{x}^\star \gets \arg\max_{(\mathbf{x}_i, y_i) \in \mathcal{D}_T} y_i$
\end{algorithmic}
\end{algorithm}

\vspace{-0.2cm}
\section{Experimental Validation}
\label{sec:experiments}

\noindent\textbf{Benchmark Circuits.} We evaluate on six analog circuits. Table~\ref{tab:benchmarks} summarizes dimensionality, objectives, and constraints. The first four circuits use in-house transistor-level testbenches in a 180nm process; the remaining two (low-dropout regulator (LDO) and charge pump) use open-source AnalogGym environments~\cite{li2024analoggym}. Together, they span low-dimensional operational amplifiers (OpAmps) with tight stability constraints (Two-stage, Three-stage), exponential-dominated temperature optimization (bandgap reference), a high-dimensional fully differential difference source-degenerated trans-conductance (FDDSD-Gm, 53 variables), and power-management macros with regime-switching behavior (LDO, charge pump). If a single pre-trained surrogate performs well across all six, it provides strong evidence that the learned function-family prior captures recurring analog patterns.

\noindent\textbf{Baseline Methods.} We compare against twenty-five optimizers. Table~\ref{tab:baselines} organizes baselines by algorithmic family. This set reflects the current toolbox for analog sizing and includes recent state-of-the-art methods alongside classical approaches.

\noindent\textbf{Evaluation Protocol.} All experiments use 10 independent random seeds. CPN initializes with five random samples to stress-test cold-start capability. Baselines follow their recommended initialization (approximately 20 points as prescribed in the original papers), placing CPN at a deliberate disadvantage to isolate the effect of structure-matched priors. Evaluation budgets are 400 SPICE simulations per run. All simulations execute on a workstation with an AMD 7950X CPU and 64GB RAM, with surrogate inference performed on an NVIDIA 2080 Ti GPU.

\begin{table}[t]
\centering
\vspace{-0.4cm}
\caption{Benchmark circuit specifications.}
\vspace{-0.4cm}
\label{tab:benchmarks}
\scriptsize
\resizebox{0.9\columnwidth}{!}{
\begin{tabular}{@{}lccl@{}}
\toprule
\textbf{Circuit} & \textbf{Vars} & \textbf{Objective} & \textbf{Key Constraints} \\
\midrule
Two-stage OpAmp     & 12 & min Current & Gain, PM, GBW \\
Three-stage OpAmp   & 18 & min Current & Gain, PM, GBW \\
Bandgap             & 20 & min TC      & Current, PSRR \\
FDDSD-Gm            & 53 & max FoM     & Gm, PM, Noise, THD \\
LDO                 & 21 & min Area    & PSRR, GBW, PM, Power \\
Charge Pump         & 36 & min Cost     & Matching, Deviation, Stability \\
\bottomrule
\end{tabular}
}
\vspace{-0.5cm}
\end{table}

\begin{table}[t]
\centering
\caption{Baseline optimizers grouped by algorithmic family.}
\vspace{-0.4cm}
\label{tab:baselines}
\scriptsize
\resizebox{0.9\columnwidth}{!}{
\begin{tabular}{@{}ll@{}}
\toprule
\textbf{Family} & \textbf{Methods} \\
\midrule
GP-based BO & \makecell[l]{Standard BO (RBF, Mat\'ern, Linear),\\Vanilla BO~\cite{gao2024vanilla}, TuRBO~\cite{turbo}, cVTS~\cite{zhao2023cvts}} \\
\midrule
\makecell[l]{High-dim.\\constrained BO} & \makecell[l]{MACE~\cite{zhang2021efficient}, REMBO~\cite{wangrembo2013}, RoSE-Opt~\cite{cao2024rose},\\SMAC~\cite{lindauer2022smac3}, USeMOC~\cite{belakaria2020uncertainty},\\tSS-BO~\cite{gu2024tss}, VGT~\cite{VGT-uai2024}} \\
\midrule
Derivative-free & \makecell[l]{BOBYQA~\cite{cartis2022BOBYQA}, DIRECT~\cite{jones1993DIRECT},\\Nelder-Mead~\cite{nelder1965Nelder–Mead}, L-BFGS-B~\cite{byrd1995L-BFGS-B}} \\
\midrule
Evolutionary & \makecell[l]{CMA-ES~\cite{hansen2016cma}, DE~\cite{rapin2018de}, PSO~\cite{vural2012pso},\\OnePlusOne~\cite{beyer2002oneplusone}, TBPSA~\cite{oquab2019tbpsa}} \\
\midrule
Portfolio & ABBO~\cite{meunier2021ABBO}, Shiwa~\cite{liu2020Shiwa}, MARIO~\cite{li2025mario} \\
\midrule
Neural surrogate & INSIGHT~\cite{poddar2025insight} \\
\midrule
Random baseline & RandomSearch \\
\bottomrule
\end{tabular}
}
\vspace{-0.5cm}
\end{table}

\begin{table*}[t]
\centering
\scriptsize
\vspace{-0.4cm}
\caption{Small-sample regression $R^2$ performance across circuits, ranging from \protect\colorbox{cbad}{\textbf{BAD}} to \protect\colorbox{chigh}{\textbf{GOOD}} performance.}
\label{tab:small_sample_regression}
\vspace{-0.3cm}
\renewcommand\arraystretch{1.2}
\setlength{\tabcolsep}{4pt}

\begin{adjustbox}{width=\textwidth,center}
\begin{tabular}{l || ccc | ccc | ccc | ccc | ccc | ccc} 
\hline
\rowcolor{white}
\textbf{} & \multicolumn{3}{c|}{\textbf{Two-stage}} & \multicolumn{3}{c|}{\textbf{Three-stage}} & \multicolumn{3}{c|}{\textbf{Bandgap}} & \multicolumn{3}{c|}{\textbf{FDDSD-Gm}} & \multicolumn{3}{c|}{\textbf{LDO}} & \multicolumn{3}{c}{\textbf{Charge Pump}} \\
\cline{2-19}
Sample size & 50 & 100 & 500 & 50 & 100 & 500 & 50 & 100 & 500 & 50 & 100 & 500 & 50 & 100 & 500 & 50 & 100 & 500 \\
\hline
\hline

GP-RBF 
& \cm 0.942 & \ch 0.970 & \ch 0.994 
& \cm 0.870 & \cm 0.931 & \ch 0.989 
& \cb -0.002 & \cl 0.163 & \cl 0.672 
& \cl 0.793 & \cm 0.916 & \cm 0.939 
& \cl 0.529 & \cl 0.637 & \cm 0.804 
& \cl 0.541 & \cl 0.679 & \cm 0.818 \\

GP-Mat\'ern 
& \cm 0.946 & \ch 0.974 & \ch 0.995 
& \cm 0.875 & \cm 0.936 & \ch 0.991 
& \cb -0.002 & \cl 0.164 & \cl 0.697 
& \cl 0.793 & \cm 0.916 & \cm 0.943 
& \cl 0.544 & \cl 0.653 & \cm 0.815 
& \cl 0.549 & \cl 0.683 & \cm 0.826 \\

GP-Linear 
& \cm 0.942 & \cm 0.946 & \cm 0.948 
& \cm 0.829 & \cm 0.906 & \cm 0.918 
& \cb -0.446 & \cb -0.090 & \cl 0.124 
& \cl 0.188 & \cm 0.897 & \cm 0.937 
& \cl 0.452 & \cl 0.592 & \cl 0.686 
& \cl 0.356 & \cl 0.673 & \cm 0.803 \\

INSIGHT 
& \cb -0.037 & \cb -0.003 & \cm 0.889 
& \cb -0.028 & \cb -0.008 & \cl 0.557 
& \cb -0.009 & \cb -0.009 & \cb -0.001 
& \cb -0.044 & \cb -0.105 & \cb -0.002 
& \cb -0.043 & \cb -0.077 & \cb -0.007 
& \cb -0.168 & \cb -0.064 & \cb -0.064 \\

\hline 

\textbf{CPN (Ours)} 
& \ch \textbf{0.988} & \ch \textbf{0.991} & \ch \textbf{0.999} 
& \ch \textbf{0.974} & \ch \textbf{0.998} & \ch \textbf{0.999} 
& \ch \textbf{0.988} & \ch \textbf{0.995} & \ch \textbf{0.999} 
& \ch \textbf{0.927} & \ch \textbf{0.958} & \ch \textbf{0.988} 
& \cm \textbf{0.843} & \cm \textbf{0.904} & \ch \textbf{0.976} 
& \cl \textbf{0.683} & \cm \textbf{0.829} & \cm \textbf{0.919} \\
\hline

\end{tabular}
\end{adjustbox}
\vspace{-0.4cm}
\end{table*}

\vspace{-0.2cm}
\subsection{Small-Sample Regression: Function-Family Priors vs Task-Specific Surrogates}
\label{sec:small_sample_regression}

We first test whether pre-trained function-family priors achieve superior regression accuracy at BO-relevant sample sizes (50--500) compared to per-circuit trained GPs and neural models. For each circuit, we construct fixed datasets with 50, 100, and 500 samples using Latin hypercube sampling, applying an 8:2 train-test split identical across all methods. Target variables are circuit-specific dominant metrics. We compare CPN against three GP variants (RBF, Mat\'ern-5/2, Linear) and INSIGHT neural surrogate, measuring test $R^2$ with all models operating on identical data partitions.

Table~\ref{tab:small_sample_regression} shows that CPN consistently achieves the highest accuracy across all circuits and sample sizes. At 50--100 samples, GP performance is highly unstable: it fits smooth OTA mappings reasonably well but collapses on circuits dominated by exponential responses or regime transitions (Bandgap, LDO, Charge Pump), often yielding near-zero or negative $R^2$. The INSIGHT also underfits in this regime: unlike CPN, which brings a pre-trained function-family prior, INSIGHT is designed to be trained from scratch on $10^3$--$10^4$ samples per topology, so when restricted to the same 50--500 samples as other methods, it behaves as an over-parameterized task-specific model and generalizes poorly. In contrast, CPN maintains uniformly strong accuracy even at 50 samples and reaches near-perfect fits by 500 samples, indicating that classical GP kernels and per-circuit neural surrogates struggle to learn highly non-stationary analog responses from scratch, whereas a function-family prior captured by CPN provides reliable, high-fidelity prediction throughout the small-data regime typical of analog BO.

\vspace{-0.2cm}
\subsection{Constrained Optimization: Structure-Awareness at Regime Boundaries}
\label{sec:constrained_time}

\begin{figure}[t]
\centering
\includegraphics[width=\columnwidth]{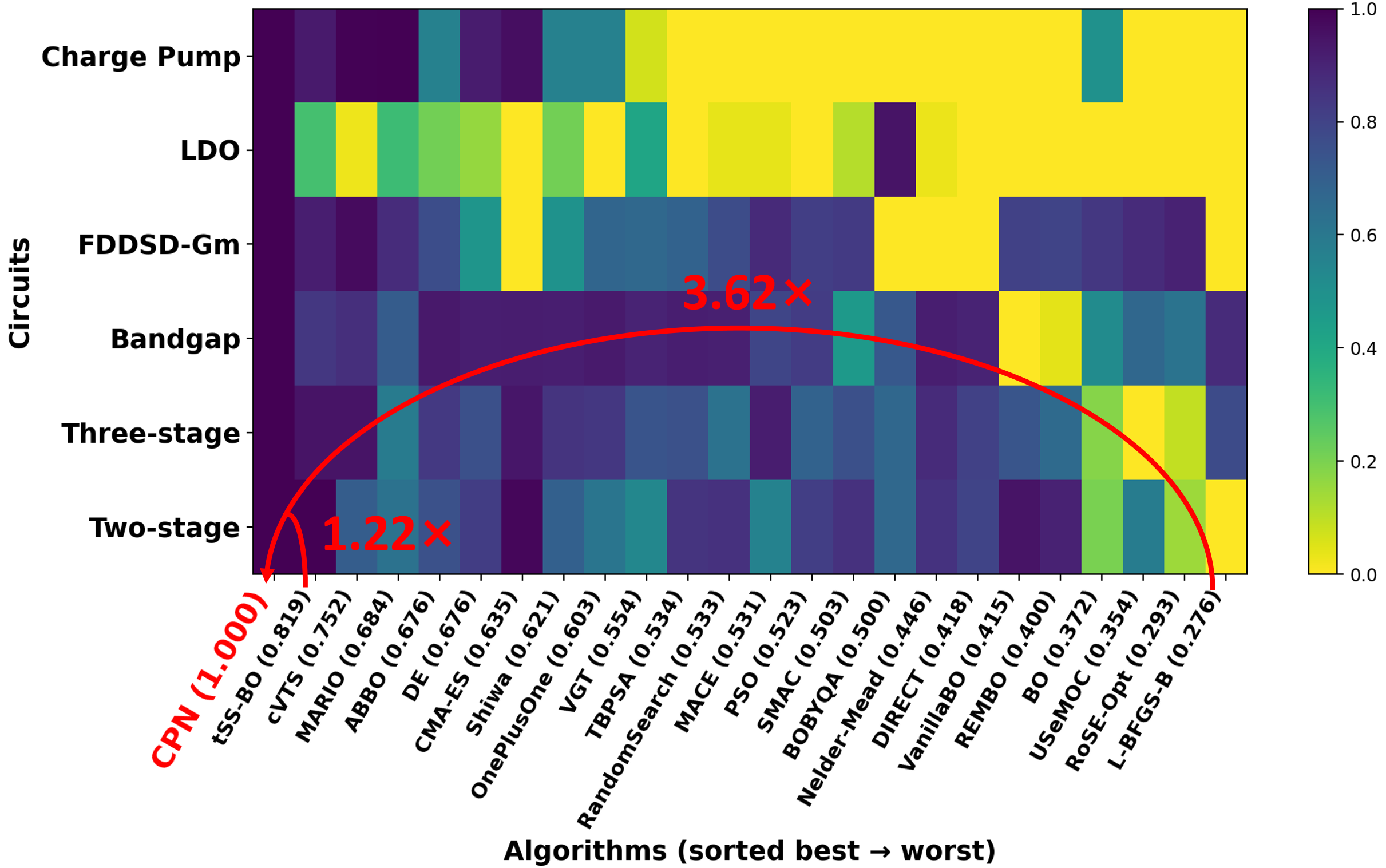}
\vspace{-0.6cm}
\caption{Normalized Performance Comparison for Constrained Optimization.}
\label{fig:constrained_heatmap}
\vspace{-0.6cm}
\end{figure}

Each circuit is treated as a constrained optimization task using the objectives and specifications in Table~\ref{tab:benchmarks}. We summarize performance by the final best constrained objective, normalized per circuit, so the best method attains a score of 1 (Figure~\ref{fig:constrained_heatmap}). To assess practical design-cycle efficiency, we also measure the end-to-end time required to reach a standardized quality threshold, including algorithmic overhead and SPICE simulation. Table~\ref{tab:time_efficiency} reports these times for representative high-performing baselines.

\begin{table}[t]
\centering
\caption{End-to-End Time (Seconds) to Reach 70\% of the Gap Between Best and Worst Performance (Methods Reaching the Target on $\geq$4 Circuits under Constraint Optimization).}
\vspace{-0.3cm}
\resizebox{\columnwidth}{!}{
\begin{tabular}{@{}lcccccc@{}}
\toprule
\textbf{Method} & \textbf{\makecell{Two-\\stage}} & \textbf{\makecell{Three-\\stage}} & \textbf{Bandgap} & \textbf{\makecell{FDDSD-\\Gm}} & \textbf{LDO} & \textbf{\makecell{Charge\\Pump}} \\
\midrule
ABBO      & 413 & 168 & 192 & 2108 & -- & -- \\
DE        & 454 & 439 & 354 & -- & -- & 1801 \\
cVTS      & 233 & 143 & 438 & 2012 & -- & 375 \\
CMA-ES    & 195 & 241 & 166 & -- & -- & 1107 \\
tSS-BO    & 72 & 222 & 157 & 5201 & -- & 15160 \\
\hline
\rowcolor{lightorange}
\textbf{CPN (Ours)}       & \textbf{47} & 199 & 167 & 2719 & \textbf{4056} & \textbf{284} \\
\bottomrule
\end{tabular}
}
\vspace{-0.6cm}
\label{tab:time_efficiency}
\end{table}

The heatmap in Figure~\ref{fig:constrained_heatmap} shows that CPN forms a consistently dark column across all six circuits. It achieves the best normalized constrained objective everywhere and improves the aggregate score by roughly 1.22$\times$ over the strongest competing optimizer and about 3.62$\times$ over the weakest one, as annotated in the figure. Table~\ref{tab:time_efficiency} indicates these quality gains do not come at the expense of runtime. On Two-stage and Charge Pump, CPN reaches the target in 47s and 284s, respectively, faster than all baselines. For LDO, only CPN reaches the target at all. These results indicate a favorable quality-time trade-off for constrained optimization compared to existing BO and evolutionary optimizers.

\vspace{-0.2cm}
\subsection{FoM Optimization: Baseline Comparisons and Performance Evaluation}

We now ask whether optimizing the FoM for analog circuit sizing can be efficiently handled using a single TabPFN v2 model, offering superior optimization performance compared to traditional surrogate-based methods such as Gaussian processes and other baseline optimizers. All methods are tasked with unconstrained FoM maximization, with a fixed budget of 400 evaluations per optimization run. We measure optimization performance through the final FoM value after 400 evaluations and assess convergence efficiency by tracking the number of iterations needed to achieve 80\% of the best FoM value or an FoM of 7.20 for the FDDSD-Gm circuit to better differentiate methods.

Table~\ref{tab:fom_itr80} presents FoM and convergence efficiency across six circuits. CPN achieves the highest FoM on all six benchmarks, with average improvements of $1.05\times$ to $3.81\times$ across different circuits. Notable results include Charge Pump, where CPN reaches FoM of 32.18 with $3.81\times$ average gain and up to $5.36\times$ improvement over the weakest baseline, and Bandgap, where CPN attains FoM of 7.70, outperforming the best baseline cVTS at 6.42 by 20\%. For convergence efficiency, CPN requires $3.34\times$ to $11.89\times$ fewer iterations on average, with maximum speedups exceeding $22\times$ on Two-stage and FDDSD-Gm, reaching target performance in 18 and 17 iterations, respectively compared to 400+ for most competitors.
This combination of superior final performance and accelerated convergence confirms that CPN, when utilizing a direct FoM prediction approach with structure-matching prior data, effectively captures the performance characteristics across various circuit types, reducing the overhead typically associated with iterative surrogate engineering.

\begin{table}[t]
\centering
\caption{Modeling strategy comparison. All methods optimize FoM; constraint-decomposed results converted via Eq.~\ref{eq:fom}. Avg. Time reports mean wall-clock per circuit over 400 iterations.}
\vspace{-0.3cm}
\label{tab:formulation_ablation}
\small
\resizebox{\columnwidth}{!}{
\begin{tabular}{lcccccc}
\toprule
\textbf{Circuit} &
\textbf{Const-Dec.} &
\textbf{Metric-Dec.} &
\textbf{Direct FoM} \\
\midrule
Two-stage       & 6.483          & 6.483          & \textbf{6.483} \\
Three-stage     & 6.956          & \textbf{6.957} & 6.952          \\
Bandgap         & 6.430          & 6.882          & \textbf{7.700} \\
FDDSD-Gm        & \textbf{7.233} & \textbf{7.233} & 7.229          \\
LDO             & 11.113         & 11.454         & \textbf{12.447} \\
Charge Pump     & 31.388         & 32.140         & \textbf{32.179} \\
\midrule
Avg. Time (s)   & 6290            & 9339            & \textbf{2174}   \\
\bottomrule
\end{tabular}
}
\vspace{-0.65cm}
\end{table}

\begin{table*}[t]
\centering
\scriptsize
\vspace{-0.4cm}
\caption{Final FoM and convergence efficiency on six circuits, ranging from \protect\colorbox{cbad}{\textbf{BAD}} to \protect\colorbox{chigh}{\textbf{GOOD}} performance.}
\vspace{-0.3cm}
\label{tab:fom_itr80}
\setlength{\tabcolsep}{3pt}

\renewcommand\arraystretch{1.2} 

\begin{adjustbox}{width=\textwidth,center}
\begin{tabular}{l | cc | cc | cc | cc | cc | cc}
\hline
\rowcolor{white}
& \multicolumn{2}{c|}{\textbf{Two-stage}} 
& \multicolumn{2}{c|}{\textbf{Three-stage}} 
& \multicolumn{2}{c|}{\textbf{Bandgap}} 
& \multicolumn{2}{c|}{\textbf{FDDSD-Gm}} 
& \multicolumn{2}{c|}{\textbf{LDO}} 
& \multicolumn{2}{c}{\textbf{Charge Pump}} \\
\cline{2-13}
\rowcolor{white}
Method & FoM & Itr@80\% & FoM & Itr@80\% & FoM & Itr@80\% & FoM & Iter@7.20 & FoM & Itr@80\% & FoM & Itr@80\% \\
\hline
\hline

RandomSearch  
& \cl 4.91 & \cb 400+ & \cl 5.57 & \cl 303   & \cl 6.28 & \cl 163  & \cl 7.21 & \cb 327 & \cl 10.0217 & \cb 400+ & \cl 6.00 & \cb 400+ \\
PSO           
& \cl 4.95 & \cb 400+ & \cl 5.23 & \cb 400+ & \cl 6.20 & \cl 283  & \cl 7.20 & \cl 66  & \cl 10.0217 & \cb 400+ & \cl 6.00 & \cb 400+ \\
OnePlusOne    
& \cm 6.47 & \cl 38   & \cl 6.31 & \cl 147  & \cl 6.27 & \cl 187  & \cl 7.16 & \cb 400+ & \cl 10.0218 & \cb 400+ & \cm 18.13 & \cb 400+ \\
DIRECT        
& \cl 5.16 & \cb 400+ & \cl 5.99 & \cl 318  & \cl 6.04 & \cb 400+ & \cb 4.63 & \cb 400+ & \cl 10.0217 & \cb 400+ & \cl 6.00 & \cb 400+ \\
DE            
& \cl 5.98 & \cl 119   & \cm 6.47 & \cl 165 & \cl 6.22 & \cl 315  & \ch 7.23 & \cl 48  & \cl 10.0234 & \cb 400+ & \cl 15.24 & \cb 400+ \\
Nelder-Mead   
& \cl 6.26 & \cl 81   & \cm 6.80 & \cl 76  & \cl 6.25 & \cl 138  & \ch 7.23 & \cm 34  & \cm 10.1536 & \cm 306  & \cl 15.11 & \cb 400+ \\
CMA-ES        
& \cl 6.33 & \cl 175  & \cm 6.76 & \cl 69  & \cl 6.28 & \cl 254  & \ch 7.23 & \cl 65  & \cl 10.0217 & \cb 400+ & \cm 18.05 & \cb 400+ \\
L-BFGS-B      
& \cb 3.85 & \cb 400+ & \cl 5.32 & \cb 400+ & \cl 6.23 & \cl 367 & \cb 4.63 & \cb 400+ & \cl 10.0217 & \cb 400+ & \cl 6.00 & \cb 400+ \\

BO            
& \cl 4.90 & \cb 400+ & \cl 5.81 & \cl 57  & \cl 5.56 & \cb 400+ & \cl 7.22 & \cl 62  & \cl 10.0217 & \cb 400+ & \cl 6.95 & \cb 400+ \\
BOBYQA        
& \cb 4.35 & \cb 400+ & \cl 5.80 & \cl 109   & \cl 5.96 & \cb 400+ & \cl 7.22 & \cl 110 & \cl 10.0280 & \cb 400+ & \cl 6.00 & \cb 400+ \\
REMBO (IJCAI'13)         
& \cl 5.41 & \cl 240  & \cl 6.38 & \ch 20  & \cb 4.91 & \cb 400+ & \cl 7.22 & \cl 53  & \cl 10.0217 & \cb 400+ & \cl 6.09 & \cb 400+ \\
TBPSA (CEC’19)         
& \cl 5.38 & \cl 334   & \cl 6.28 & \cl 232  & \cl 6.23 & \cl 237  & \cl 7.22 & \cl 164 & \cl 10.0217 & \cb 400+ & \cl 6.00 & \cb 400+ \\
USeMOC (AAAI'20)        
& \cl 4.90 & \cb 400+ & \cl 5.67 & \cl 39  & \cl 5.79 & \cb 400+ & \cl 7.22 & \cm 34  & \cl 10.0217 & \cb 400+ & \cl 6.00  & \cb 400+ \\

Shiwa (GECCO'20)         
& \cl 6.22 & \cl 130  & \cm 6.49 & \cm 25  & \cl 6.28 & \cl 100  & \cl 7.16 & \cb 400+ & \cl 10.0241 & \cb 400+ & \cm 18.13 & \cb 400+ \\
ABBO (TEC'21)           
& \cl 6.28 & \cl 132  & \cm 6.49 & \cm 25  & \cl 6.25 & \cl 104  & \cl 7.16 & \cb 400+ & \cl 10.0241 & \cb 400+ & \cm 18.13 & \cb 400+ \\
MACE (TCAD'21)          
& \cl 6.35 & \cl 41   & \cm 6.80 & \cl 33  & \cl 5.99 & \cb 400+ & \cl 7.22 & \cl 38  & \cl 10.0227 & \cb 400+ & \cl 6.00  & \cb 400+ \\
TuRBO (DAC'21)         
& \cl 6.23 & \cm 35   & \cm 6.61 & \cl 66  & \cl 6.25 & \cl 156  & \cl 7.21 & \cm 24  & \cl 10.0302 & \cb 400+ & \cm 23.05 & \cb 400+ \\
SMAC (JMLR'22)          
& \cl 5.45 & \cl 89   & \cm 6.54 & \cl 86  & \cl 5.46 & \cb 400+ & \cl 7.22 & \cl 89  & \cl 10.0246 & \cb 400+ & \cl 6.00  & \cb 400+ \\
cVTS (DAC'23)          
& \cl 6.34 & \cl 90   & \cm 6.94 & \cl 51  & \cl 6.42 & \cl 130  & \cl 7.22 & \cl 39  & \cl 10.0218 & \cb 400+ & \cm 28.74 & \cl 228  \\
VGT (UAI'24)           
& \cl 5.40 & \cm 24   & \cm 6.50 & \cm 25  & \cl 6.20 & \cl 161 & \cl 7.21 & \cm 23  & \cl 10.0363 & \cb 400+ & \cl 6.00  & \cb 400+ \\
RoSE-Opt (TCAD'24)      
& \cl 4.87 & \cb 400+ & \cl 5.73 & \ch 13  & \cl 5.69 & \cb 400+ & \cl 7.22 & \cm 34  & \cl 10.0217 & \cb 400+ & \cl 6.00  & \cb 400+ \\
Vanilla BO (ICCAD'24)     
& \cl 6.35 & \cl 149  & \cm 6.51 & \ch 11  & \cb 4.87 & \cb 400+ & \cl 7.22 & \cl 113 & \cl 10.0217 & \cb 400+ & \cl 6.00  & \cb 400+ \\
tSS-BO (DATE'24)        
& \cm 6.47 & \cl 93   & \cm 6.87 & \cl 81  & \cl 6.08 & \cb 400+ & \ch 7.23 & \cm 26  & \cl 10.0308 & \cb 400+ & \cl 14.18 & \cb 400+ \\
MARIO (DAC'25)         
& \cl 5.62 & \cl 165   & \cl 6.28 & \cl 110  & \cl 6.23 & \cl 281  & \ch 7.23 & \cm 29  & \cm 10.1644 & \cm 288  & \cm 29.66 & \cm 183  \\
\hline

\textbf{CPN (Ours)}     
& \ch \textbf{6.48} & \ch \textbf{18}   
& \ch \textbf{6.95} & \ch \textbf{23}  
& \ch \textbf{7.70} & \ch \textbf{73}  
& \ch \textbf{7.23} & \ch \textbf{17}  
& \ch \textbf{12.4474} & \ch \textbf{77}   
& \ch \textbf{32.18} & \ch \textbf{115}  \\

\textbf{Avg Imprv.} 
& \ch \textbf{1.18}$\times$ & \ch \textbf{11.89}$\times$ 
& \ch \textbf{1.12}$\times$ & \ch \textbf{5.18}$\times$ 
& \ch \textbf{1.29}$\times$ & \ch \textbf{3.92}$\times$ 
& \ch \textbf{1.05}$\times$ & \ch \textbf{8.28}$\times$ 
& \ch \textbf{1.24}$\times$ & \ch \textbf{5.08}$\times$ 
& \ch \textbf{3.81}$\times$ & \ch \textbf{3.34}$\times$ \\

\textbf{Max Imprv.} 
& \ch \textbf{1.68}$\times$ & \ch \textbf{22.22}$\times$ 
& \ch \textbf{1.33}$\times$ & \ch \textbf{17.39}$\times$ 
& \ch \textbf{1.58}$\times$ & \ch \textbf{5.48}$\times$ 
& \ch \textbf{1.56}$\times$ & \ch \textbf{23.53}$\times$ 
& \ch \textbf{1.24}$\times$ & \ch \textbf{5.19}$\times$ 
& \ch \textbf{5.36}$\times$ & \ch \textbf{3.48}$\times$ \\
\hline
\end{tabular}
\end{adjustbox}
\vspace{-0.4cm}
\end{table*}

\subsection{Modeling Strategy Comparison}
\label{sec:formulation_ablation}

With CPN's surrogate and acquisition function fixed, the modeling strategy affects both optimization quality and computational efficiency. We compare the three strategies from Section~\ref{sec:cpn_formulations} under identical conditions: 400-evaluation budgets, same initialization, and DEI acquisition. All variants optimize the same FoM objective, with constraint-decomposed results converted to FoM via Eq.~(\ref{eq:fom}) for unified comparison.

Table~\ref{tab:formulation_ablation} reports the final best FoM and average wall-clock time per circuit. Direct FoM modeling achieves best or tied-best performance on 5 of 6 circuits. On Bandgap, it reaches FoM of 7.700 versus 6.882 (metric-decomposed) and 6.430 (constraint-decomposed)—a 12\% improvement over the next-best strategy and 20\% over constraint-decomposed. On LDO, direct modeling attains 12.447 compared to 11.454 and 11.113, gains of 8.7\% and 12.0\%, respectively. Metric-decomposed modeling performs competitively on Three-stage (6.957) and FDDSD-Gm (7.233) but degrades substantially on Bandgap and LDO, suggesting sensitivity to metric coupling structure.

Computational efficiency favors direct modeling decisively. It completes optimization 2.9$\times$ faster than constraint-decomposed (2174s vs 6290s) and 4.3$\times$ faster than metric-decomposed (9339s). This speedup stems directly from surrogate query cost: direct modeling evaluates one PFN per candidate, while alternatives evaluate $N_c+1$ or $N_m$ instances. On circuits with 6+ specifications, this overhead compounds across thousands of acquisition evaluations per iteration.

The performance pattern reveals a data concentration effect. Circuits with strong exponential sensitivities (Bandgap temperature coefficient, LDO transient response) benefit most from concentrating all observations into learning the composite objective rather than fragmenting them across coupled metrics. Smoother rational-function circuits (Three-stage, FDDSD-Gm) show smaller differences, as metric coupling is less critical. Combined with its computational advantage, direct FoM modeling provides the recommended default strategy for CPN deployment on new circuits.

\vspace{-0.2cm}
\subsection{Ablation Study: Acquisition Impact}
\label{sec:ei_ablation}

\begin{table}[t]
\centering
\caption{Acquisition ablation showing normalized improvement factor (EI/DEI for minimization, DEI/EI for maximization). Values $>1$ indicate DEI outperforms EI.}
\vspace{-0.3cm}
\label{tab:ablation_acquisition}
\scriptsize
\resizebox{\columnwidth}{!}{
\begin{tabular}{@{}lcc@{}}
\hline
\textbf{Circuit} & \textbf{Factor @ 30 iters} & \textbf{Factor @ 50 iters} \\
\hline
Two-stage   & \textbf{3.33} & \textbf{3.33} \\
Three-stage & 0.99 & 0.89 \\
Bandgap     & \textbf{1.59} & 0.97 \\
FDDSD-Gm    & 0.99 & 1.00 \\
LDO         & 1.00 & \textbf{1.08} \\
Charge Pump & \textbf{1.03} & 1.00 \\
\hline
\end{tabular}
}
\vspace{-0.5cm}
\end{table}

With the surrogate fixed, we expect a distribution-consistent acquisition such as DEI to be most helpful in the small-sample regime where posteriors are strongly non-Gaussian. We therefore keep the CPN surrogate and constrained objectives fixed and only swap the acquisition between EI and DEI. On each of the six circuits, we run constrained optimization with identical initialization and record the best task objective after 30 and 50 simulations; these checkpoints deliberately target the high-uncertainty phase, where Section~\ref{sec:small_sample_regression} shows that CPN still has $R^2<0.9$ on LDO and Charge Pump within the first 50 samples. For OTA and bandgap/LDO/charge-pump circuits, this objective is minimized (e.g., current), while for FDDSD-Gm it is maximized (FoM). To obtain a compact comparison, we convert the two objectives into a single improvement factor in favor of DEI, reported as “EI/DEI” in Table~\ref{tab:ablation_acquisition}.

At 30 evaluations, DEI reduces the objective on Two-stage by about 70\% (improvement factor $\approx 3.3\times$) and on Bandgap by roughly 37\%, and still yields a small improvement on Charge Pump. By 50 evaluations, as the surrogate fit improves, the gap shrinks on most circuits and the factor moves closer to 1, but DEI still dominates on Two-stage and improves LDO by about 7\% while never being substantially worse than EI (factors remain essentially 1 on FDDSD-Gm and within a few percent of 1 on the remaining benchmarks). Combined with the regression results, this supports our claim that DEI is particularly useful when the PFN posterior is still inaccurate and non-Gaussian, and that it offers a robust default choice that does not hurt performance once the model has converged.

\section{Conclusion}
\label{sec:conclusion}
We introduced CPN, the first framework to view analog sizing as inference over function-family priors derived from circuit primitives. This perspective connects circuit theory with foundation-model inference, enabling a single reusable prior across heterogeneous designs without kernel tuning or task-specific surrogate work. Across six benchmarks and twenty-five baselines, CPN achieves higher small-sample accuracy, better constrained solutions, and faster convergence. Ablations show that direct FoM supervision and DEI further strengthen performance in low-data regimes. These results suggest a shift from hand-crafted GP kernels toward physics-grounded foundation models for scalable analog automation.




\bibliographystyle{ACM-Reference-Format}


\bibliography{CPN_reference}

@inproceedings{poddar2025insight,
  title={INSIGHT: A Universal Neural Simulator Framework for Analog Circuits with Autoregressive Transformers},
  author={Poddar, Souradip and Oh, Youngmin and Lai, Yao and Zhu, Hanqing and Hwang, Bosun and Pan, David Z},
  booktitle={2025 62nd ACM/IEEE Design Automation Conference (DAC)},
  pages={1--7},
  year={2025},
  organization={IEEE}
}

@article{zhang2021efficient,
  title={An efficient batch-constrained bayesian optimization approach for analog circuit synthesis via multiobjective acquisition ensemble},
  author={Zhang, Shuhan and Yang, Fan and Yan, Changhao and Zhou, Dian and Zeng, Xuan},
  journal={IEEE Transactions on Computer-Aided Design of Integrated Circuits and Systems},
  volume={41},
  number={1},
  pages={1--14},
  year={2021},
  publisher={IEEE}
}

@inproceedings{xing2024kato,
  title={KATO: Knowledge alignment and transfer for transistor sizing of different design and technology},
  author={Xing, Wei W and Fan, Weijian and Liu, Zhuohua and Yao, Yuan and Hu, Yuanqi},
  booktitle={Proceedings of the 61st ACM/IEEE Design Automation Conference},
  pages={1--6},
  year={2024}
}

@article{hollmann2025accurate,
  title={Accurate predictions on small data with a tabular foundation model},
  author={Hollmann, Noah and M{\"u}ller, Samuel and Purucker, Lennart and Krishnakumar, Arjun and K{\"o}rfer, Max and Hoo, Shi Bin and Schirrmeister, Robin Tibor and Hutter, Frank},
  journal={Nature},
  volume={637},
  number={8045},
  pages={319--326},
  year={2025},
  publisher={Nature Publishing Group UK London}
}

@inproceedings{li2025mario,
  title={MARIO: A Superadditive Multi-Algorithm Interworking Optimization Framework for Analog Circuit Sizing},
  author={Li, Wangzhen and Meng, Yuan and Lyu, Ruiyu and Yan, Changhao and Zhu, Keren and Bi, Zhaori and Zhou, Dian and Zeng, Xuan},
  booktitle={2025 62nd ACM/IEEE Design Automation Conference (DAC)},
  pages={1--7},
  year={2025},
  organization={IEEE}
}

@inproceedings{li2024analoggym,
  title={AnalogGym: an open and practical testing suite for analog circuit synthesis},
  author={Li, Jintao and Zhi, Haochang and Lyu, Ruiyu and Li, Wangzhen and Bi, Zhaori and Zhu, Keren and Zeng, Yanhan and Shan, Weiwei and Yan, Changhao and Yang, Fan and others},
  booktitle={Proceedings of the 43rd IEEE/ACM International Conference on Computer-Aided Design},
  pages={1--9},
  year={2024}
}

@InProceedings{relu,
  title = 	 {Deep Sparse Rectifier Neural Networks},
  author = 	 {Glorot, Xavier and Bordes, Antoine and Bengio, Yoshua},
  booktitle = 	 {Proceedings of the Fourteenth International Conference on Artificial Intelligence and Statistics},
  pages = 	 {315--323},
  year = 	 {2011},
  editor = 	 {Gordon, Geoffrey and Dunson, David and Dudík, Miroslav},
  volume = 	 {15},
  series = 	 {Proceedings of Machine Learning Research},
  address = 	 {Fort Lauderdale, FL, USA},
  month = 	 {11--13 Apr},
  publisher =    {PMLR},
  url = 	 {https://proceedings.mlr.press/v15/glorot11a.html}
}

@inproceedings{belakaria2020uncertainty,
  author       = {Syrine Belakaria and
                  Aryan Deshwal and
                  Nitthilan Kannappan Jayakodi and
                  Janardhan Rao Doppa},
  title        = {Uncertainty-Aware Search Framework for Multi-Objective Bayesian Optimization},
  booktitle    = {Proc. AAAI},
  pages        = {10044--10052},
  publisher    = {{AAAI} Press},
  year         = {2020},
  bibsource    = {dblp computer science bibliography, https://dblp.org}
}

@inproceedings{gu2024tss,
  title={tSS-BO: Scalable Bayesian Optimization for Analog Circuit Sizing via Truncated Subspace Sampling},
  author={Gu, Tianchen and Wang, Jiaqi and Bi, Zhaori and Yan, Changhao and Yang, Fan and Qin, Yajie and Cui, Tao and Zeng, Xuan},
  booktitle={2024 Design, Automation \& Test in Europe Conference \& Exhibition (DATE)},
  pages={1--6},
  year={2024},
  organization={IEEE}
}

@inproceedings{
VGT-uai2024,
title={Exploring High-dimensional Search Space via Voronoi Graph Traversing},
author={Zhao, Aidong and Zhao, Xuyang and Gu, Tianchen and Bi, Zhaori and Sun, Xinwei and Yan, Changhao and Yang, Fan and Zhou, Dian and Zeng, Xuan},
booktitle={The 40th Conference on Uncertainty in Artificial Intelligence},
year={2024}
}

@article{lindauer2022smac3,
  title={SMAC3: A versatile Bayesian optimization package for hyperparameter optimization},
  author={Lindauer, Marius and Eggensperger, Katharina and Feurer, Matthias and Biedenkapp, Andr{\'e} and Deng, Difan and Benjamins, Carolin and Ruhkopf, Tim and Sass, Ren{\'e} and Hutter, Frank},
  journal={The Journal of Machine Learning Research},
  volume={23},
  number={1},
  pages={2475--2483},
  year={2022},
  publisher={JMLRORG}
}

@inproceedings{turbo,
  title={Local bayesian optimization for analog circuit sizing},
  author={Touloupas, Konstantinos and Chouridis, Nikos and Sotiriadis, Paul P},
  booktitle={Proc. DAC},
  pages={1237--1242},
  publisher  = {{IEEE}},
  address = {San Francisco, California, USA},
  year={2021},
  organization={IEEE}
}

@inproceedings{zhao2023cvts,
  title={cvts: A constrained voronoi tree search method for high dimensional analog circuit synthesis},
  author={Zhao, Aidong and Wang, Xianan and Lin, Zixiao and Bi, Zhaori and Li, Xudong and Yan, Changhao and Yang, Fan and Shang, Li and Zhou, Dian and Zeng, Xuan},
  booktitle={2023 60th ACM/IEEE Design Automation Conference (DAC)},
  pages={1--6},
  year={2023},
  organization={IEEE}
}

@inproceedings{wangrembo2013,
author = {Wang, Ziyu and Zoghi, Masrour and Hutter, Frank and Matheson, David and De Freitas, Nando},
title = {Bayesian optimization in high dimensions via random embeddings},
year = {2013},
isbn = {9781577356332},
publisher = {AAAI Press},
booktitle = {Proceedings of the Twenty-Third International Joint Conference on Artificial Intelligence},
pages = {1778–1784},
numpages = {7},
location = {Beijing, China},
series = {IJCAI '13}
}

@article{cao2024rose,
  title={Rose-opt: Robust and efficient analog circuit parameter optimization with knowledge-infused reinforcement learning},
  author={Cao, Weidong and Gao, Jian and Ma, Tianrui and Ma, Rui and Benosman, Mouhacine and Zhang, Xuan},
  journal={IEEE Transactions on Computer-Aided Design of Integrated Circuits and Systems},
  year={2024},
  publisher={IEEE}
}

@inproceedings{gao2024vanilla,
  title={Is Vanilla Bayesian Optimization Enough for High-Dimensional Architecture Design Optimization?},
  author={Gao, Yuanhang and Luo, Donger and Bai, Chen and Yu, Bei and Geng, Hao and Sun, Qi and Zhuo, Cheng},
  booktitle={Proceedings of the 43rd IEEE/ACM International Conference on Computer-Aided Design},
  pages={1--9},
  year={2024}
}

@inproceedings{
    muller2022transformers,
    title={Transformers Can Do Bayesian Inference},
    author={Samuel M{\"u}ller and Noah Hollmann and Sebastian Pineda Arango and Josif Grabocka and Frank Hutter},
    booktitle={International Conference on Learning Representations},
    year={2022}
}

@article{hansen2016cma,
  title={The CMA evolution strategy: A tutorial},
  author={Hansen, Nikolaus},
  journal={arXiv preprint arXiv:1604.00772},
  year={2016}
}

@misc{rapin2018de,
  title={Nevergrad-A gradient-free optimization platform},
  author={Rapin, J{\'e}r{\'e}my and Teytaud, Olivier},
  year={2018}
}

@inproceedings{oquab2019tbpsa,
  title={Parallel noisy optimization in front of simulators: optimism, pessimism, repetitions, population control},
  author={Oquab, Maxime and Rapin, Jeremy and Teytaud, Olivier and Cazenave, Tristan},
  booktitle={Workshop Data-driven Optimization and Applications at CEC},
  year={2019}
}

@article{cartis2022BOBYQA,
  title={Escaping local minima with local derivative-free methods: a numerical investigation},
  author={Cartis, Coralia and Roberts, Lindon and Sheridan-Methven, Oliver},
  journal={Optimization},
  volume={71},
  number={8},
  pages={2343--2373},
  year={2022},
  publisher={Taylor \& Francis}
}

@article{meunier2021ABBO,
  title={Black-box optimization revisited: Improving algorithm selection wizards through massive benchmarking},
  author={Meunier, Laurent and Rakotoarison, Herilalaina and Wong, Pak Kan and Roziere, Baptiste and Rapin, J{\'e}r{\'e}my and Teytaud, Olivier and Moreau, Antoine and Doerr, Carola},
  journal={IEEE Transactions on Evolutionary Computation},
  volume={26},
  number={3},
  pages={490--500},
  year={2021},
  publisher={IEEE}
}

@inproceedings{liu2020Shiwa,
  title={Versatile black-box optimization},
  author={Liu, Jialin and Moreau, Antoine and Preuss, Mike and Rapin, Jeremy and Roziere, Baptiste and Teytaud, Fabien and Teytaud, Olivier},
  booktitle={Proceedings of the 2020 Genetic and Evolutionary Computation Conference},
  pages={620--628},
  year={2020}
}

@article{vural2012pso,
  title={Analog circuit sizing via swarm intelligence},
  author={Vural, R Acar and Yildirim, T{\"u}lay},
  journal={AEU-International journal of electronics and communications},
  volume={66},
  number={9},
  pages={732--740},
  year={2012},
  publisher={Elsevier}
}

@article{jones1993DIRECT,
  title={Lipschitzian optimization without the Lipschitz constant},
  author={Jones, Donald R and Perttunen, Cary D and Stuckman, Bruce E},
  journal={Journal of optimization Theory and Applications},
  volume={79},
  number={1},
  pages={157--181},
  year={1993},
  publisher={Springer}
}

@article{nelder1965Nelder–Mead,
  title={A simplex method for function minimization},
  author={Nelder, John A and Mead, Roger},
  journal={The computer journal},
  volume={7},
  number={4},
  pages={308--313},
  year={1965},
  publisher={The British Computer Society}
}

@article{byrd1995L-BFGS-B,
  title={A limited memory algorithm for bound constrained optimization},
  author={Byrd, Richard H and Lu, Peihuang and Nocedal, Jorge and Zhu, Ciyou},
  journal={SIAM Journal on scientific computing},
  volume={16},
  number={5},
  pages={1190--1208},
  year={1995},
  publisher={SIAM}
}

@article{beyer2002oneplusone,
  title={Evolution strategies--a comprehensive introduction},
  author={Beyer, Hans-Georg and Schwefel, Hans-Paul},
  journal={Natural computing},
  volume={1},
  number={1},
  pages={3--52},
  year={2002},
  publisher={Springer}
}

@article{enz1995analytical,
  title={An analytical MOS transistor model valid in all regions of operation and dedicated to low-voltage and low-current applications},
  author={Enz, Christian C and Krummenacher, Fran{\c{c}}ois and Vittoz, Eric A},
  journal={Analog integrated circuits and signal processing},
  volume={8},
  number={1},
  pages={83--114},
  year={1995},
  publisher={Springer}
}

@book{gray2009analysis,
  title={Analysis and design of analog integrated circuits},
  author={Gray, Paul R and Hurst, Paul J and Lewis, Stephen H and Meyer, Robert G},
  year={2009},
  publisher={John Wiley \& Sons}
}

\end{document}